\documentclass[11pt,a4paper]{article}
\usepackage[dvipsnames]{xcolor}
\usepackage[hyperref]{acl2020}

\usepackage{times}
\usepackage{latexsym}
\usepackage{framed}
\usepackage{enumitem}
\usepackage{subcaption}
\usepackage{graphicx}
\usepackage{booktabs}
\usepackage{multirow}
\usepackage{multicol}
\usepackage{enumitem}

\usepackage{xspace}
\usepackage{amsmath}
\usepackage{amssymb}
\usepackage{microtype}

\usepackage{tcolorbox}

\usepackage{soul}

\aclfinalcopy % Uncomment this line for the final submission
\title{\emph{I like fish \includegraphics[height=12pt]{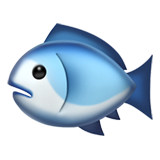}, especially dolphins \includegraphics[height=12pt]{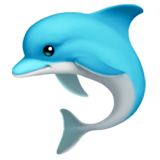}:}\thanks{* Dolphins are mammals, not fish.}  \\ Addressing Contradictions in Dialogue Modeling}
      
\author{Yixin Nie$^1$, Mary Williamson$^2$, Mohit Bansal$^1$, Douwe Kiela$^2$, Jason Weston$^2$\\
  $^1$UNC Chapel Hill\\$^2$Facebook AI Research\\}

\date{}

\begin{document}
\maketitle
\begin{abstract}
To quantify how well natural language understanding models can capture consistency in a general conversation, we introduce the DialoguE COntradiction DEtection task (DECODE) and a new conversational dataset containing both human-human and human-bot contradictory dialogues. We then compare a structured utterance-based approach of using pre-trained Transformer models for contradiction detection with the typical unstructured approach. Results reveal that: (i) our newly collected dataset is notably more effective at providing supervision for the dialogue contradiction detection task than existing NLI data including those aimed to cover the dialogue domain; (ii) the structured utterance-based approach is more robust and transferable on both analysis and out-of-distribution dialogues than its unstructured counterpart.  We also show that our best contradiction detection model correlates well with human judgements and further provide evidence for its usage in both automatically evaluating % (as a metric)
and improving the consistency of state-of-the-art generative chatbots.
\end{abstract}

\section{Introduction}

% large scale conversational data.
% success in neural response generation and its usage in open-domain chat bots.

Recent progress on neural approaches to natural language processing~\cite{devlin2019bert, brown2020gpt3}, and the availability of large amounts of conversational data~\cite{lowe-etal-2015-ubuntu, smith-etal-2020-put} have triggered a resurgent interest on building intelligent open-domain chatbots.
% Advancements have been made on building intelligent open-domain chatbots due to the recent progress on neural approaches to natural language processing~\cite{devlin2019bert, brown2020gpt3} and the availability of large amounts of conversational data.
Newly developed end-to-end neural bots~\cite{zhang-etal-2020-dialogpt, adiwardana2020towards, roller2020recipes} 
%have advanced the field, as measured by 
are claimed to be superior to their predecessors~\cite{worsnick2018mitsuku, zhou2020design} using 
various human evaluation techniques~\cite{see-etal-2019-makes, li2019acute, adiwardana2020towards} that aim to give a more accurate measure of what makes a good conversation. While the success is indisputable, there is still a long way to go before we arrive at human-like open-domain chatbots. For example, it has been shown that open-domain chatbots frequently generate annoying errors~\cite{adiwardana2020towards, roller2020recipes} and a notorious one among these is the class of contradiction, or consistency, errors.

\begin{figure}[t!]
    \centering
    \includegraphics[width=\columnwidth]{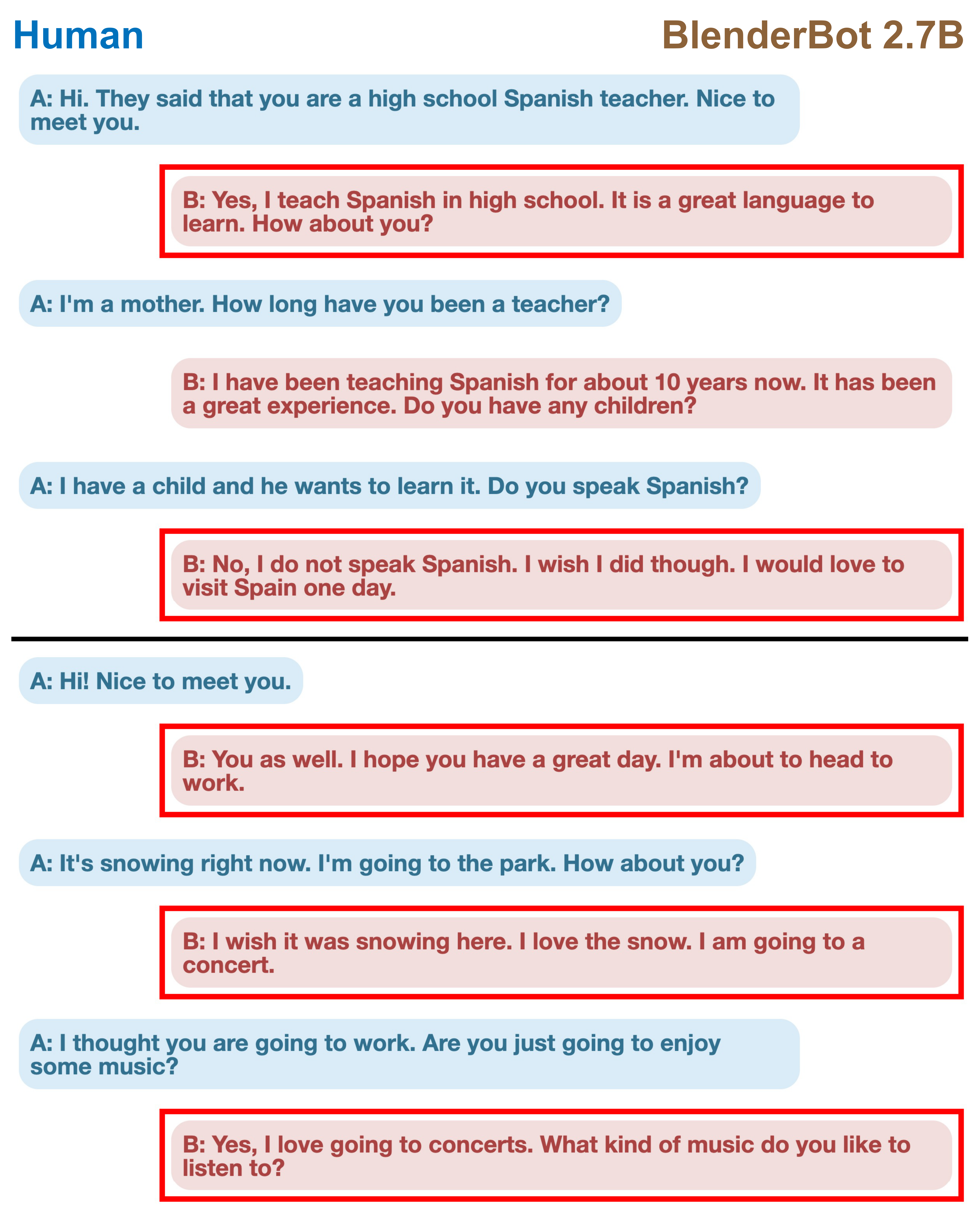}
    \caption{Two dialogue examples demonstrating a state-of-the-art chatbot (B)~\cite{roller2020recipes} contradicting itself when talking to a human (A).}
    \label{fig:contradiction_example}
\end{figure}

When interacting with chatbots, people carry over many of the same expectations as when interacting with humans~\cite{nass2000machines}. 
Self-contradictions (see examples in \autoref{fig:contradiction_example}) by these bots are often jarring, immediately disrupt the conversational flow, and help support arguments about whether generative models could ever really understand what they are saying at all \cite{marcus2018deep}. 
From a listener's perspective, such inconsistent bots fail to gain user trust and their long-term communication confidence.
From a speaker's perspective, it violates the maxim of quality in the Grice's cooperative principle~\cite{grice1975logic} ---"Do not say what you believe to be false." Hence, efforts on reducing contradicting or inconsistent conversations by open-domain chatbots are imperative.

Historically, modularizing dialogue systems, i.e., assigning an aspect of conversational modeling to a specific component and then integrating it back into the dialogue system, can often help improve overall system satisfaction~\cite{fang2017sounding, chen2018gunrock}. Prior works~\cite{welleck-etal-2019-dialogue} characterized the modeling of persona-related consistency as a natural language inference (NLI) problem~\cite{dagan2005pascal, bowman-etal-2015-large}, constructed a dialog NLI dataset based on Persona-Chat~\cite{zhang-etal-2018-personalizing}, but so far state-of-the-art chatbots \cite{roller2020recipes} have not been able to make use of such techniques.
%and tried to reduce inconsistency in generative models via unlikelihood training on contradicting utterances~\cite{li-etal-2020-dont}, with preliminary results based on perplexity only.
Overall, the challenge remains that we are still unable to answer the simple yet important question---``\textit{how well can a natural language understanding module model the consistency (including persona, logic, causality, etc) in a general conversation?}". The lack of an ability to measure this obscures to what degree building new modules or techniques can in turn help prevent contradicting responses during generation.

Seeking to answer this question, we introduce the DialoguE COntradiction DEtection task (DECODE)\footnote{Our DECODE dataset is publicly available at \url{https://parl.ai/projects/contradiction}.} and collect a new conversational dataset containing human written dialogues where one of the speakers deliberately contradicts what they have previously said at a certain point during the conversation. 
We also collect an out-of-distribution (OOD) set of dialogues in human-bot interactive settings which contain human-labeled self-contradictions made by different chatbots.

We then compare a set of state-of-the-art systems, including a standard unstructured approach and a proposed structured approach for utilizing NLI models to detect contradictions. In the unstructured approach, a Transformer NLI model directly takes in the concatenation of all utterances of the input dialogue for prediction, following the paradigm of NLU modeling. In the structured approach, utterances are paired separately before being fed into Transformer NLI models, explicitly taking account of the natural dialogue structure.

Results reveal that: (1) our newly collected dataset is notably more effective at providing supervision for the contradiction detection task than existing NLI data including those aimed at covering the dialogue domain; (2) the structured utterance-based approach for dialogue consistency modeling is more robust in our analysis and more transferable to OOD human-bot conversation than the unstructured approach. This finding challenges the mainstream unstructured approach of simply applying  pre-trained Transformer models and expecting them to learn the structure, especially for OOD scenarios which are often the case when incorporating NLU modules into NLG systems, since intermediate in-domain data are scarce. 

Finally, with such improvements on the contradiction detection task, we show that our best resultant contradiction detector correlates well with human judgements and can be suitable for use as an automatic metric for checking dialogue consistency. We further provide evidence for its usage in improving the consistency of state-of-the-art generative chatbots.

\section{Related Work}
Several prior works on improving dialogue consistency have explored using direct modeling of the dialogue context in generation algorithms. The modeling can be implicit where the dialogue consistency-related information like style~\cite{wang-etal-2017-steering}, topics, or personal facts are maintained in distributed embeddings~\cite{li-etal-2016-persona, zhang2019neural}, neural long-term memories~\cite{bang2015example}, hierarchical neural architecture~\cite{serban2016building}, latent variables~\cite{serban2017hierarchical}, topical attention~\cite{dziri2019augmenting}, or even self-learned feature vectors~\cite{zhang2019consistent}.
Some works have grounded generation models on explicit user input~\cite{qian2018assigning}, or designated personas~\cite{zhang-etal-2018-personalizing}. Although, improvements on automatic generation metrics were often shown on guided response generation based on the consistency modeling, the issue of contradiction has never been resolved, nor have generally applicable methods to gauge the consistency improvements been developed. Further, simply scaling models has not made the problem go away, as is evident in the largest chatbots trained such as BlenderBot with up to 9.4B parameter Transformers \cite{roller2020recipes}.
%It is often not flexible enough to have an intermediate score quantifying the consistency improvement in those methods.

More similar to our work is utilizing NLI models in dialogue consistency. ~\citet{dziri-etal-2019-evaluating-coherence} attempted to use entailment models trained on synthetic datasets for dialogue topic coherence evaluation. Particularly, ~\citet{welleck-etal-2019-dialogue} constructed the dialogue NLI dataset and \cite{li-etal-2020-dont} utilized it to try to reduce inconsistency in generative models via unlikelihood training in a preliminary study that reports perplexity results, but did not measure actual generations or contradiction rates. We note that the dialogue NLI dataset is only semi-automatically generated, with limited coverage of only persona-chat data \cite{zhang-etal-2018-personalizing}, whereas our DECODE is human-written and across diverse domains. Our task also involves logical and context-related reasoning beyond personal facts, for example the dialogue at the bottom of \autoref{fig:contradiction_example} shows a non-persona-related contradiction. We show in our experiments that transfer of DECODE is subsequently more robust than dialogue NLI on both human-human and human-bot chats. 

% we also show? depend on results

% of In mostly of the direct consistency modeling, improvement have been shown on automatic generation metrics.

% Dataset statistics
% 
\section{Task and Data}
% The task formalization.
\subsection{Dialogue Contradiction Detection}
\label{sec:task_dialogue_contradiction_detection}
% To quantify the modeling of dialogue consistency, 
We formalize dialogue contradiction detection as a supervised classification task. The input of the task is a list of utterances $x=\{u_0, u_1, u_2, ..., u_n\}$ representing a dialogue or a dialogue snippet. The output is $y$, indicating whether the last utterance $u_n$ contradicts any previously conversed information contained in the dialogue$\{u_0, u_1, u_2, ..., u_{n-1}\}$, where $y$ can be $0$ or $1$ corresponding to the non-contradiction and the contradiction label respectively. Preferably, the output should also include a set of indices $\mathbf{I}\subseteq \{0, 1, ..., n-1\}$ representing a subset of $\{u_0, u_1, u_2, ..., u_{n-1}\}$ which contain information that is actually contradicted by the last utterance $u_{n}$. The extra indices $\mathbf{I}$ output require models to pinpoint the evidence for the contradiction, providing an extra layer of explainability. 
% The collection is centered around the formalization.

\subsection{Data Collection}

\begin{figure*}[t]
    \centering
    \includegraphics[width=\textwidth]{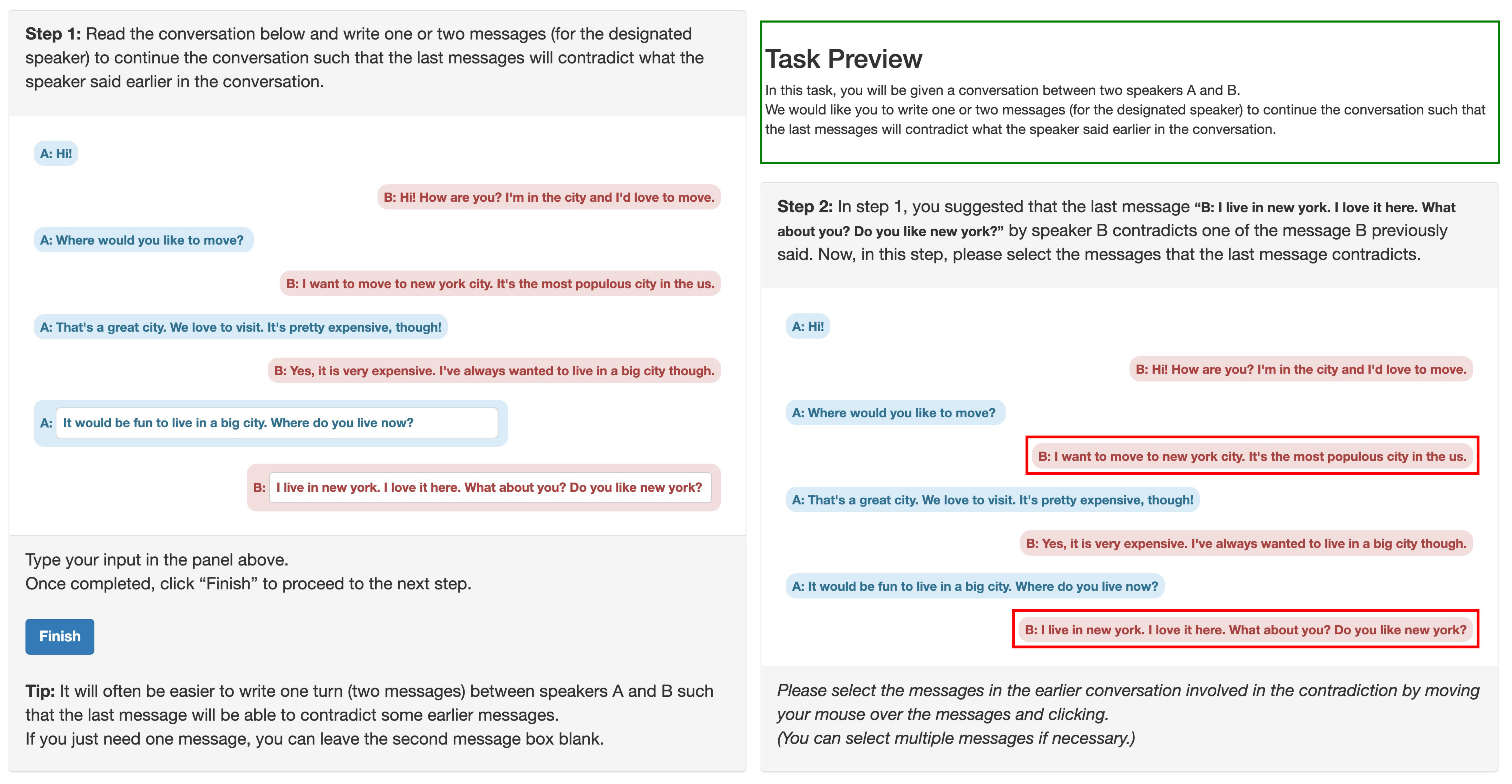}
    \caption{The collection interface. The task preview box (top right) gives a short description of the task before the annotator will work on the writing. The collection consists of two steps. In Step 1 (on the left), the annotators are asked to write one or two utterances such that the last utterance will contradict some previous utterances in the conversation. In Step 2 (on the right), the annotators are asked to pick the utterances in the conversation that are involved in the contradiction. We use a casual term ``message" instead of ``utterance'' in the instructions.}
    \label{fig:collection_interface}
\end{figure*}

\paragraph{Annotation Design}
Our goal is first to collect training and evaluation data for this task. We thus collect dialogues in which the last utterance contradicts some previous utterances in the dialogue history. To obtain such dialogues, we give annotators dialogue snippets from pre-selected dialogue corpora, and then ask them to continue the conversation by writing one or two utterances such that the last utterance by the last speaker contradicts the dialogue history. We also ask annotators to mark all the utterances in the dialogue history that are involved in the contradiction as supporting evidence. \autoref{fig:collection_interface} shows the annotation user interface. We ask annotators to write contradicting utterances based partly on existing dialogues rather than collecting new dialogue from scratch because the provided dialogues can often convey semantic-rich contexts from different domains and inspire annotators to write more diverse examples. We crowdsource the continuation and annotation data with Amazon Mechanical Turk and the collection is based on the ParlAI\footnote{https://parl.ai~\cite{miller2017parlai}} framework.

\paragraph{Quality Control}
% Quality Control
We apply the following mechanism to ensure the quality of collected data:
\begin{itemize}[leftmargin=1em]
\vspace{-6pt}
\setlength\itemsep{-0.1em}
    \item \textbf{Onboarding Test}: Every annotator needs to pass an onboarding test before they can actually contribute dialogue examples. The test is the same dialogue contradiction detection task as in the actual collection procedure, including 5 dialogues where 3 of them have an ending utterance that contradicts the dialogue history. The annotator needs to select the correct label (contradiction or non-contradiction) for all five dialogues to pass the test. This mechanism tests whether an annotator understands the task.
    \item \textbf{Maximum Annotation Count Limit}: The maximum number of examples one annotator can create is 20. This mechanism helps further diversify the dialogue examples by reducing similar patterns that appear in one or a group of annotators~\cite{geva-etal-2019-modeling}.
    \item \textbf{Verification}: This subtask ensures that the dialogue examples indeed contain an ending utterance that contradicts the dialogue history. We ask 3 additional annotators to verify some of the collected examples and select the ones where all three verifiers agreed on the contradiction label, and use these for our resulting validation and tests sets. This mechanism ensures that there is a clear, agreed-upon contradiction in the dialogue, preventing the subjectivity and ambiguity issues in some NLU tasks~\cite{nie-etal-2020-learn}. See the appendix for statistics about the data verification.
\end{itemize}

% additional annotator/annotation statistics can be shown here.
% time of annotation, number of example per annotator, etc...

% \subsection{Data Collection}
\subsection{Dataset}
\label{sec:dataset}
We collected 17,713 human-written contradicting dialogues in which 4,121 are verified by 3 annotators. The pre-selected dialogue source corpora are Wizard of Wikipedia ~\cite{dinan2018wizard}, \textsc{EmpatheticDialogues} ~\cite{rashkin-etal-2019-towards}, Blended Skill Talk ~\cite{smith-etal-2020-put}, and ConvAI2~\cite{dinan2020second}, covering various conversational topics. To facilitate the evaluation of consistency modeling on the dialogue contradiction detection classification task, we sample an equal number of non-contradicting dialogues according to the same dialogue length distribution as the contradicting ones from the same dialogue corpus.\footnote{We balance the labels in the dataset following the standard NLI evaluation~\cite{bowman-etal-2015-large, welleck-etal-2019-dialogue}.} Then, we make the split such that the train split contains unverified examples, and dev and test splits only contain verified examples. Each split has balanced labels between contradiction and non-contradiction dialogues. \autoref{tab:dataset_source_statistics} shows the  breakdown of each of the dataset sources and data splits.

\begin{table}[t]
    \centering
     \small
    \scalebox{1}{
    \begin{tabular}{lrrr}
    \toprule
     & \textbf{Train} & \textbf{Dev} & \textbf{Test}\\
    \midrule
    Wizard of Wikipedia & 6,234 & 1,208 & 1,160 \\
     \textsc{EmpatheticDialogues} & 6,182 & 1,046 & 1,050 \\
    Blended Skill Talk & 8,554 & 1,200 & 1,310 \\
    ConvAI2 & 6,214 & 572 & 696\\
    \textbf{Total} & \textbf{27,184} & \textbf{4,026} & \textbf{4,216}\\
    \bottomrule
    \end{tabular}
    }
    \caption{Our DECODE Main Dataset source statistics. The labels in each split are balanced. There are a total of 2,013$+$2,108 contradicting examples in the dev and test sets which are the collected 4,121 verified examples. The first column indicates the source of the dialogue.}
    \label{tab:dataset_source_statistics}
\end{table}

% \subsubsection{}
\noindent
\textbf{Auxiliary (Checklist) Test Sets } 
We further create two auxiliary checklist evaluation sets by transforming the contradiction examples in the original test in two ways such that the ground truth label is either invariant or expected to change. The two resultant sets serve as diagnostic tests on the behavior, generalization and transferability of our models.

The transformations are described below:
\begin{itemize}[leftmargin=1em]
\vspace{-6pt}
\setlength\itemsep{-0.1em}
\item \textbf{Add Two Turns (A2T)} We insert a pair of randomly sampled utterances into the dialogue such that the inserted utterances are between the two original contradicting utterances. This gives a new contradicting dialogue with a longer dialogue history.
\item \textbf{Remove Contradicting Turns (RCT)} We remove all the turns (all pairs of utterances)\footnote{All the dialogues in the dataset involved two speakers that takes turns in speaking. To maintain this structure, for each marked utterance we remove a pair of utterance that represents a turn of conversation. This also helps remove the information that was involved in the contradiction such that the resultant label should be ``non-contradiction".} marked as supporting evidence for the contradiction in the dialogue except the last utterance. This results in a new non-contradiction dialogue.
\end{itemize}

Notice that the two data transformations we used were based on utterance-level evidence annotations and therefore are not applicable for DNLI and other NLI data.

\begin{table}[t]
    \centering
     \small
    \scalebox{1}{
    \begin{tabular}{lrc}
    \toprule
     & \textbf{Count} & \textbf{Label}\\
    \midrule
    Main (Train) & 27,184 & balanced \\
    Main (Dev) & 4,026 & balanced \\
    Main (Test) & 4,216 & balanced \\
    \midrule
    Human-Bot (Test) & 764 & balanced \\
    \midrule
    A2T (Test) & 2,079 & contradiction \\
    RCT (Test) & 2,011 & non-contradiction \\
    \bottomrule
    \end{tabular}
    }
    \caption{DECODE Dataset summary. The first column presents the different  dataset types. ``Main" is the collected human-written dialogues. ``balanced" indicates that the contradiction and non-contradiction labels in that part of the dataset are balanced. A2T and RCT are the auxiliary test sets described in Sec. \ref{sec:dataset}.}.
    \label{tab:dataset_statistics}
\end{table}

\noindent
\textbf{Human-Bot Test Set }
% The Q-func data
Our main collected dataset involves human-written dialogues containing contradicting utterances based on human-human dialogues from existing corpora. In practice, to evaluate the response quality of a machine rather than a human in terms of its consistent responses, we care about how well a contradiction detector can perform in human-bot interactive conversations. To that end, we further collect human-bot dialogue data by employing workers on Amazon Mechanical Turk to interact with a diverse set of open-domain bots. These include Poly-encoder~\cite{humeau2019poly} based retrieval models,
generative models~\cite{roller2020recipes}, unlikelihood trained models~\cite{li2019don}, retrieve-and-refine models~\cite{weston2018retrieve,roller2020recipes}, models either pre-trained on
  a previously existing Reddit dataset extracted and obtained by a third party that was hosted by pushshift.io \cite{baumgartner2020pushshift} or fine-tuned on the Blended Skill Talk (BST) dialogue tasks \cite{smith2020can} -- that is, all the dialogue models that are compared in the study in \cite{roller2020recipes}.
During the collection, if the bot generates an utterance that contradicts itself, we ask the worker to mark the utterance. In some of the dialogues, workers are explicitly instructed to goad the bots into making contradicting utterances. The final human-bot test set we derive contains 764 dialogues, half of which ends with a contradicting utterance by the bot. All the dialogues in the set, with either contradiction or non-contradiction labels, are verified by 3 additional annotators, beside the human who actually talked to the bot.

The auxiliary and human-bot test sets are aimed to test models' robustness and generalizability beyond accuracy on the collected human-written test set~\cite{ribeiro-etal-2020-beyond, gardner-etal-2020-evaluating}, and give a more comprehensive analysis of the task. \autoref{tab:dataset_statistics} summarizes the final overall dataset. \autoref{tab:examples} gives one example for each dataset type.

\begin{table}[t]
\tiny
\scalebox{1}{
	\begin{tabular}{p{34em}}
		\toprule
		\textbf{Main (Human-Human)}\\
		\textcolor{OliveGreen}{Human1: \textit{I recently had a long weekend with some old friends. It was fantastic.}}\\ 
		Human2: \textit{It must have been fun to catch up with them.}\\
		\textcolor{OliveGreen}{Human1: \textit{it was. We've spent a lot of time together and apart now, so it was good to catchup.}}\\ 
		  Human2: \textit{Well I hope you guys continue to stay in touch.}\\
		\textcolor{OliveGreen}{Human1: \textit{Sometimes friends grow apart over time though.}}\\ 
		  Human2: \textit{Yeah, that's true. You should forget about them and go on with your life.}\\
        \midrule
        \textbf{Human-Bot}:\\
        \textcolor{OliveGreen}{Human1: \textit{It's a beautiful day. I'm going to take a walk later.}}\\ 
		  Bot: \textit{What do you do for a living?}\\
		\textcolor{OliveGreen}{Human1: \textit{I work in a doctor's office. How about yourself?}}\\ 
		  Bot: \textit{I am a stay at home mom.}\\
		\textcolor{OliveGreen}{Human1: \textit{That's great! How many kids do you have?}}\\ 
		  Bot: \textit{I don't have any kids yet, but I plan on getting married soon.}\\
        \midrule
        \textbf{Add Two Turns (A2T)}:\\
        \textcolor{OliveGreen}{Human1: \textit{i hate when ash from my cigarette drops in my work pants}}\\ 
		  Human2: \textit{oof that sucks really bad}\\
		\textcolor{OliveGreen}{Human1: \textit{yeah, i haave to wait till i get home to get the stain off, it is really embarrasing}}\\ 
		  Human2: \textit{yea i can imagine it is}\\
		\underline{Human1: \textit{Every time I look at it I remember the good times we had together.}}\\ 
% 		I honestly hope to.
		  \underline{Human2: \textit{well thats nice}}\\
		 \textcolor{OliveGreen}{Human1: \textit{I will have to wash the stain with soap and water.}}\\ 
		  Human2: \textit{Ash stains on your pants is not a big deal though.}\\
        \midrule
        \textbf{Remove Contradicting Turns (RCT)}:\\
        \textcolor{OliveGreen}{Human1: \textit{I was disgusted when I noticed the food on the table}}\\ 
		  Human2: \textit{What kind of food?}\\
		\textcolor{OliveGreen}{\st{Human1: \textit{It was brussel sprouts and Liver}}}\\ 
		  \st{Human2: \textit{Oh, disgusting.}}\\
		\textcolor{OliveGreen}{Human1: \textit{I couldn't even bear to take a single bite}}\\ 
		  Human2: \textit{Brussel sprouts and liver sounds delicious to me!}\\
	  \bottomrule
	\end{tabular}}
	\caption{Dialogue examples for different dataset types. Underline indicates that the pair of utterances is randomly added. Strikethrough text indicates that the pair of utterances is removed. Dialogue examples for Human-Human, Human-Bot, and A2T end with a contradicting utterance whereas the example for RCT has an ending utterance whereby the original contradicting pair of utterances in the dialogue history are removed.\label{tab:examples}}
\end{table}

% show examples here... for all the category.

\section{Models}

To model the dialogue consistency task, we first employ some of the techniques used in NLI sequence-to-label modeling, where the input is a pair of textual sequences and the output is a label. The benefit of such modeling is that we can directly make use of existing NLI datasets during training. However, unlike previous work~\cite{welleck-etal-2019-dialogue} that directly utilized NLI models giving a 3-way output among ``entailment", ``contradiction", and ``neutral", we modify the model with a 2-way output between ``contradiction" and ``non-contradiction" labels. This is because the task is, in its essence, centered around the detection of inconsistency. 

More formally, we denote the model as $\hat{y}_{pred} = f_{\theta}(\mathbf{C}, u)$, where $\hat{y}_{pred}$ is the prediction of the label $y$, i.e. whether the textual response $u$ contradicts some textual context $\mathbf{C}$, and where $\theta$ are the parameters of the model. We then explore two different approaches to utilize $f_{\theta}$ for dialogue contradiction detection.

\subsection{Dialogue Contradiction Detectors}
As described in \autoref{sec:task_dialogue_contradiction_detection}, a detector is asked to determine whether the last utterance of the dialogue $u_n$ contradicts the previous dialogue history $\{u_0, u_1, u_2, ..., u_{n-1}\}$. In what follows, we describe two approaches that propose differing $f_{\theta}$ for the detection prediction problem.

\paragraph{Unstructured Approach.} In this approach, we simply concatenate all the previous utterances in the dialogue history to form a single textual context. Then, we apply $f_\theta$ to the context and the last utterance to infer the probability of contradiction.
\begin{equation}
 \hat{y}_{pred} = f_{\theta}([u_0, u_1, u_2, ..., u_{n-1}], u_n)
\end{equation}
When concatenating the utterances, we insert special tokens before each utterance to indicate the speaker of that utterance. This is aimed to provide a signal of the dialogue structure to the models. Still, this approach assumes that the model can use these features adequately to learn the underlying structure of the dialogue implicitly during training.

\paragraph{Structured Utterance-based Approach.} Since the reasoning crucially depends on the last utterance, in this method we first choose all the utterances by the last speaker to form a  set $\mathbf{S}$. We then pair every utterance in the set with the last utterance and feed them one by one into $f^{UB}_\theta$. The final contradiction probability is the maximum over all the outputs.
\begin{equation}
 \hat{y}_{pred} = \max \left\{f^{UB}_\theta(u_i, u_n) : u_i \in \mathbf{S}\right\}
\end{equation}
Additionally, the utterance-based approach is able to give a set of utterances as supporting evidence for a contradiction decision by choosing the pairs having contradiction probability higher than a threshold $\eta_{e}$:
\begin{equation}\label{eq:support}
\mathbf{I} = \left\{ i : f^{UB}_\theta(u_i, u_n) > \eta_{e}\right\}
\end{equation}
This not only gives explanations for its prediction but can also help diagnose the model itself, e.g. we can measure metrics of the model's ability to provide these explanations by comparing them against gold supporting evidence annotations from DECODE. 

One downside of this modeling approach is that it will not be able to capture reasoning between speakers. A case for that would be a pronoun by one speaker might refer to something initiated by the other speaker. 
%The utterance-based approach aims to give an initial trial on methods that utilize explicit inductive structure bias and we hope future work can explore more dedicated development on injecting dialogue structure into language modeling.
Nevertheless, the utterance-based approach explicitly adds an inductive structure bias to learning and inference which we will see can aid its generalization capability.

\paragraph{Thresholding.}
For both the unstructured and utterance-based approaches, the detection of contradiction is made by comparing $\hat{y}_{pred}$ with a threshold $\tau$ and by default $\tau$ is 0.5.

\begin{table*}[t!]
    \centering
    % \small
    \scalebox{0.84}{
    \begin{tabular}{l|l|cccc}
    
    \hline
\textbf{\small Pre-trained Model} & \textbf{\small Training Data} & \textbf{\small Main (Test)} & \textbf{\small Main (Test-Strict)} & \textbf{\small Human-Bot} & \textbf{\small SE (Precision / Recall / F1)} \\
% Majority & - & 50.00 & 50.00 & 50.00 & 50.43 / 47.10 / 48.71 \\
\hline
\midrule
\multicolumn{6}{l}{\textit{ Unstructured Approach}}\\
\hline
\multirow{7}{*}{RoBERTa}    & All & \textbf{97.46}  & - & 77.09 & - \\
                            & All - DNLI & 97.44  & - & 73.17 & - \\
                            & All - ANLI-R3 & 98.04  & - & 73.56 & - \\
                            & All - DECODE & 84.42  & - & 61.91 & - \\
                            \cline{2-6}
                            & DNLI           & 57.19  & -                & 60.34     & -               \\
                            & ANLI-R3       & 82.21  & -                & 59.69     & -               \\
                            & DECODE         & 96.85  & -                & 70.03     & -               \\
                            % & DECODE (One Speaker)         & 96.68  & -                & 73.17     & -               \\
\hline
\midrule
\multicolumn{6}{l}{\textit{ Utterance-based Approach}}\\
\hline
\multirow{7}{*}{RoBERTa} & SNLI + MNLI & 77.40 & 47.70 & 73.17 & 63.3 / 84.6 / 72.4
\\
 & All            & 94.19  & 80.08           & 83.64     & 85.9 / \textbf{91.2} / \textbf{88.5}                            \\
                                  & All - DNLI     & 94.38  & \textbf{80.93}           & 81.68     & 86.7 / 90.1 / 88.4                           \\
                                  & All - ANLI-R3 & 94.07  & 79.32           & 82.85     & 85.2 / 91.8 / 88.4                           \\
                                  & All - DECODE   & 86.67  & 66.95           & 77.36     & 78.0 / 83.4 / 80.6                           \\
                                  \cline{2-6}
                                  & DNLI           & 76.54  & 63.09           & 75.26     & 85.1 / 61.2 / 71.2                            \\
                                  & ANLI-R3       & 81.59  & 69.11           & 70.52     & \textbf{88.2} / 64.3 / 74.3                           \\
                                  & DECODE         & 93.19  & 80.86           & \textbf{84.69}     & 87.9 / 87.2 / 87.5                           \\
                                  \hline
                                  BERT                     & DECODE         & 88.88  & 74.14           & 75.52     & 84.9 / 83.7 / 84.3                            \\
                                  \hline
                                  Electra                  & DECODE         & 93.17  & 81.19           & 80.76     & 87.9 / 87.1 / 87.5                           \\
                                  \hline
                                  BART                     & DECODE         & 94.47  & 80.10            & 79.19     & 85.8 / 90.7 / 88.2                          \\
                                  \hline
                                  \midrule
\multicolumn{6}{l}{\textit{ Majority}}\\
\hline
- & - & 50.00 & 50.00 & 50.00 & 50.4 / 47.1 / 48.7 \\
\hline
\end{tabular}
    }
    \caption{Test performance of different models and approaches. ``All" in the ``Training Data" column stands for a combination of SNLI, MNLI, DNLI, ANLI-R3, DECODE. ``All - DNLI" denotes all the datasets with DNLI removed. ``SE" stands for supporting evidence. The ``Main (Test-Strict)" column indicates the performance where both the 2-way contradiction detection and the supporting evidence retrieval exactly match with the ground truth.}
    \label{tab:main_results}
\end{table*}

\subsection{Experimental Setup}
We study four base pre-trained models variants for $f_{\theta}$: BERT~\cite{devlin2019bert}, Electra~\cite{clark2019electra}, RoBERTa~\cite{liu2019roberta}, and BART~\cite{lewis-etal-2020-bart}. They represent the start-of-the-art language representation models and have yielded successes in many NLU tasks. The input format of $f_\theta$ follows how these models handle sequence-pairs ($\mathbf{C}$ and $u$) classification task with padding, separator and other special tokens such as position embeddings and segment features inserted at designated locations accordingly. 

We fine-tune $f_{\theta}$ on different combinations of NLI training data including SNLI~\cite{bowman-etal-2015-large}, MNLI~\cite{williams-etal-2018-broad}, ANLI-R3~\cite{nie-etal-2020-adversarial}\footnote{ANLI data is collected in three rounds resulting in three subsets (R1, R2, R3). We only used training data in R3 since it contains some dialogue-related examples.}, DNLI~\cite{welleck-etal-2019-dialogue}, as well as our DECODE Main training set. We convert the 3-way labels of the examples in existing NLI datasets to 2-way\footnote{The  3-way ``entailment" and ``neutral" label is converted to ``non-contradiction'' while 3-way ``contradiction" is kept the same.} and $\theta$ is optimized using cross-entropy loss. When training $f_\theta^{UB}$ in the utterance-based approach using the DECODE training set, the input sequences are sampled utterance pairs from the DECODE dialogue. 
%In the unstructured approach $f_\theta$ is trained directly using the pair of concatenation of previous utterances and the last utterances in the dialogue.
In other scenarios,  $f_\theta$ or $f_\theta^{UB}$ are trained with data treated as in normal NLI training.

The models are evaluated on the test sets described in Sec. \ref{sec:dataset}. For the utterance-based approach, which additionally provides supporting evidence utterances (\autoref{eq:support}), 
we report Precision, Recall, and F1 on these evidence predictions. We also report a stricter score which evaluates whether both 2-way contradiction detection and supporting evidence retrieval \emph{exactly match} with the ground truth on our DECODE Main test set.

% which will be updated by optimizing cross-entropy loss on differentiating contradicting and non-contradicting input pairs.

% \begin{equation}
    % p(y \given \mathbf{C}, u) = f_{\theta}(\mathbf{C}, u)
    
% \end{equation}
% where $y\in\{0, 1\}$. $y=1$ indicates that the textual response $u$ contradicts some textual context $\mathbf{C}$. The model $f_{\theta}$ outputs the probability for such a contradiction.

% We convert the 3-way labels of the examples in existing NLI datasets to 2-way for training.

% utterance-based model and unstructured model

\section{Results and Analysis}
\subsection{Performance on Constructed Dataset}
We test different pre-trained models with both the unstructured and the structured utterance-based approaches. We explicitly investigate the model performance when trained on DNLI or ANLI-R3 and compare it with DECODE because these are recently published NLI datasets that contain examples in a dialogue setting. However, we do also provide results comparing to other NLI datasets as well as multi-tasking all datasets at once, in addition to various  ablations. The results are shown in \autoref{tab:main_results}. We now describe our key observations.

\paragraph{DECODE is notably more effective than other existing NLI data in providing supervision for contradiction detection in dialogue.} 
We found that models trained on DECODE achieve higher accuracy than that of those trained on DNLI or ANLI-R3, on all evaluation sets in both the unstructured and utterance-based approach. 
On the DECODE Main test set, the utterance-based RoBERTa model trained (fine-tuned) on DECODE achieves 93.19\% accuracy, which is a 12-point jump from the same model training on ANLI-R3 and a 16-point jump from training on DNLI. The best model on human-bot data is utterance-based RoBERTa trained on DECODE with 84.69\%, while the same model trained on DNLI can only get 75.26\% accuracy, and ANLI-R3 is even worse with 70.52\%. 
While training on ``All'' datasets (SNLI, MNLI, ANLI-R3, DNLI \& DECODE) is effective, the removal of DECODE from the training data induces a consequential downgrade on the performance on all evaluation sets. 
In particular, removing DECODE training data for unstructured RoBERTa causes a 15-point loss of accuracy on the human-bot data from (77.09\% to 61.91\%). Further, training on DECODE is also more helpful than  DNLI or ANLI-R3 for supporting evidence retrieval.
These findings indicate that existing NLI data has limited transferability to the dialogue contradiction detection task despite their coverage of the dialogue domain in addition to other domains. Training on NLI data which does not cover examples with dialogue structures, e.g., SNLI+MNLI is even worse, only achieving 77.4\% on DECODE Main (Test) vs. 93.19\% for DECODE and cannot even reach the majority baseline on the ``Main (Test-Strict)".
Hence overall, this empirically demonstrates that our DECODE data provides a valuable resource for modeling dialogue consistency and developing data-driven approaches for contradiction detection.

% Data matters DNLI ANLI-R3 despite the fact that 

\paragraph{Different pre-training models that perform similarly on the in-domain test set can have very different performance on OOD human-bot dialogue.} The last four rows of the table show the results of utterance-based RoBERTa, BERT, Electra, and BART trained on DECODE. We can see that RoBERTa, Electra, and BART got similar in-domain accuracy on DECODE, around 93\%-94\%. RoBERTa stands out when comparing their performance on the human-bot test set with the highest score of 84.69\%  across the column (compared to 75.52, 79.19 and 80.76 for the other methods) and with better performance on supporting evidence retrieval as well. We speculate that this is due to the fact that RoBERTa pre-training data has a broader coverage than Electra and BART. We hope future work on dialogue contradiction detection could explore pre-training models on more dialogue-focused corpora. 

% pre-training data might effective NLU in dialogue setting.
% inspiring pre-training targetted at dialogue

% DECODE is important
% Different Models
% Unstructured vs. Utterance-based
\paragraph{The unstructured approach gets higher accuracy on the in-domain test set.}
A direct comparison between unstructured RoBERTa and utterance-based RoBERTa trained on DECODE reveals that the unstructured approach more often than not gets a higher accuracy than its corresponding utterance-based approach when other experiential setups are kept identical. Noticeably, unstructured RoBERTa trained on all NLI data got a 97.46\% score, whereas utterance-based yielded 94.19\%. This seemingly indicates that training an unstructured model is able to yield a good representation of the consistency of the dialogue. However, further analysis on the human-bot and auxiliary test sets shows that such high accuracy is an over-amplification of the model's real understanding ability, as we discuss next.

\begin{figure}[ht]
    \centering
    \includegraphics[width=\columnwidth, trim=0 0 20 20, clip]{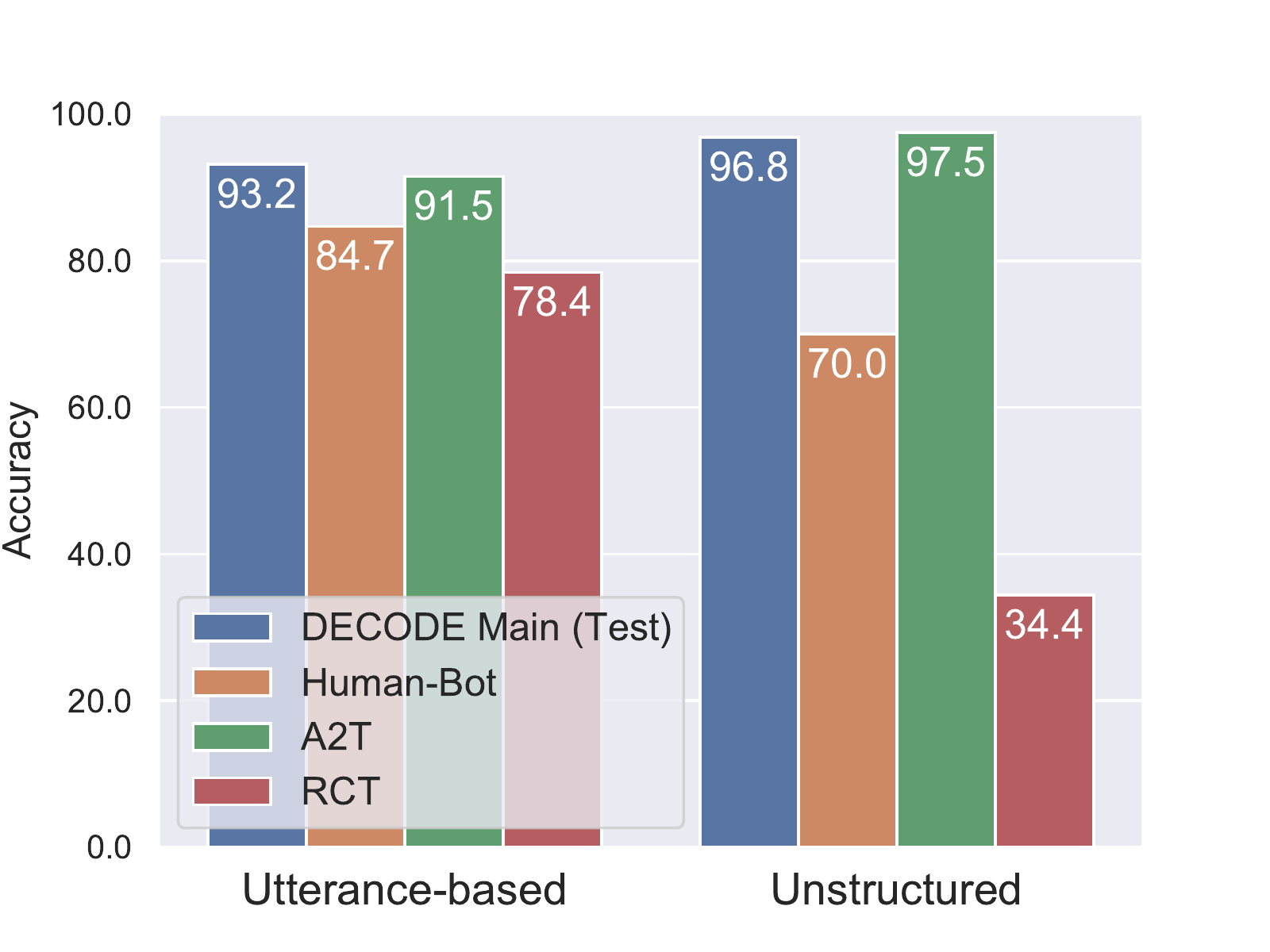}
    \caption{Comparison between utterance-based and unstructured approaches of RoBERTa pre-trained, DECODE fine-tuned models on DECODE Main (Test), Human-bot, and auxiliary test sets.}
    \label{fig:auxiliary_test_barplot}
\end{figure}

\paragraph{The structured utterance-based approach is more robust, and more transferable.}
\autoref{fig:auxiliary_test_barplot} gives a comparison between utterance-based and unstructured RoBERTa on each of the evaluation sets. We can see that the utterance-based model is able to maintain satisfactory performance across all the sets whereas the unstructured model underperforms at the human-bot and RCT auxiliary test sets with a 34.4\% accuracy on RCT compared to 78.4\% for utterance-based, in stark contrast to the high performance of the unstructured method on the in-domain DECODE Main test set. This result indicates the unstructured approach overfits on superficial patterns in the DECODE Main training data which are still present due to RCT's construction process.\footnote{Overfitting on superficial patterns is a typical issue and open problem in NLU modeling~\cite{nie-etal-2020-adversarial}.}
% here we need to borrow some citations from NLU task on overfitting superficial patterns.
The fact that the utterance-based approach has good transferability to the OOD human-bot test set indicates that injecting the correct inductive structure bias is beneficial for modeling dialogue consistency. We believe this is an interesting result generally for research using Transformers, where there is currently a belief amongst some practitioners that they can just use a standard Transformer and it will learn all the structure correctly on its own. In our setting that is not the case, and we provide a method that can rectify that failing.

\paragraph{In general, there is still much room for improvement.}
The results in \autoref{tab:main_results} also demonstrate that the modeling of dialogue consistency is a demanding task. On the contradiction detection task, the best score achieved by the state-of-the-art pre-trained language models on DECODE (Test-Strict) is 80.86\% and the best human-bot test score is 84.69\%. Considering all the examples in the test sets are verified by at least 3 annotators, humans are able to swiftly identify such contradictions. This suggests there is a large ability gap between our best automatic detectors and humans. Closing this gap is an important challenge for the community. 

% making use the automatic in actually consistenty dialogue will require more effort on modeling dialogue consistency and improvement on resolving the contradiction detection task.
% maybe put into disscussion and conclusion.

\subsection{Performance in an Interactive Setting}
\label{sec:interactive}
The results discussed above evaluate models on constructed datasets with intentionally balanced labels. This facilitates the comparison between models following a NLU evaluation perspective. In practice, we would like to evaluate how well a model can detect contradicting utterances sampled naturally from interactive human-bot dialogue. To that end, we test our trained detection models on the raw interactive human-bot dialogue data\footnote{This is the same set of dialogues from which we constructed the balanced human-bot test set.} having a total number of 764 dialogues consisting of 8,933 utterances. Since the contradiction task in naturally sampled dialogue can be extremely unbalanced, the total number of contradicting utterances in the raw dialogue list is only 381\footnote{The majority baseline accuracy is 95.73\%.}. We apply our contradiction detectors on every bot-generated utterance and calculate the precision, recall, and F1 on contradiction detection. Since the scores might be subjective to the threshold $\tau$, we also evaluate the threshold-invariant Area Under the ROC Curve (AUC)~\cite{bradley1997auc}. 

As shown in \autoref{tab:utterance_results}, model precision on the task is not satisfactory (23.94 at best). However, the best model achieves acceptable scores on both Recall and AUC. This indicates its potential usage for strict blocking of inconsistent utterances of a generative model (bot). The table also draws the same conclusion as \autoref{tab:main_results} that the structured utterance-based RoBERTa model trained using DECODE data is the best method for contradiction detection, comparing to training on other NLI data or using an unstructured approach.
In the following sections we thus use that best method as our detector for further experiments.% 

\begin{table}[t]
    \centering
    % \small
    \scalebox{0.88}{
    \begin{tabular}{l|cccc}
    
    \hline
\textbf{\small Training Data} & \textbf{\small Precision} & \textbf{\small Recall} & \textbf{\small F1} & \textbf{\small AUC} \\
% Majority & - & 50.00 & 50.00 & 50.00 & 50.43 / 47.10 / 48.71 \\
\hline
\midrule
\multicolumn{5}{l}{\textit{ Unstructured Approach}}\\
\hline
All & 15.89 & 60.11 & 25.14 & 80.47 \\
% All - DNLI & 97.44  & - & 73.17 & - \\
% All - ANLI-R3 & 98.04  & - & 73.56 & - \\
All - DECODE & 15.63 & 57.74 & 24.60 & 71.82 \\
                            
% DNLI           & 57.19  & -                & 60.34     & -               \\
% ANLI-R3       & 82.21  & -                & 59.69     & -               \\
DECODE & 17.05 & 50.13 & 25.45 & 73.40 \\
\hline
\midrule
\multicolumn{5}{l}{\textit{ Utterance-based Approach}}\\
\hline
 All    & 23.35 & 71.65	& 35.23 & 84.96 \\
% All - DNLI     & 94.38  & 80.93           & 81.68     & -        \\
% All - ANLI-R3 & 94.07  & 79.32           & 82.85     & - \\
All - DECODE   & 17.17 & 68.50 & 27.46 & 80.09 \\

DNLI & 16.32 & 65.09 & 26.09 & 79.29 \\
ANLI-R3 & 22.52 & 41.73	& 29.26 & 76.36 \\
DECODE & \textbf{23.94} & \textbf{74.28} & \textbf{36.21} & \textbf{87.16} \\
\hline
\end{tabular}
    }
    \caption{RoBERTa performance on all the bot-generated utterances from the raw interactive human-bot dialogue. The threshold $\tau$ for prediction is 0.5.}
    \label{tab:utterance_results}
\end{table}

\begin{figure}[ht]
    \centering
    \includegraphics[width=\columnwidth, trim=30 20 30 30, clip]{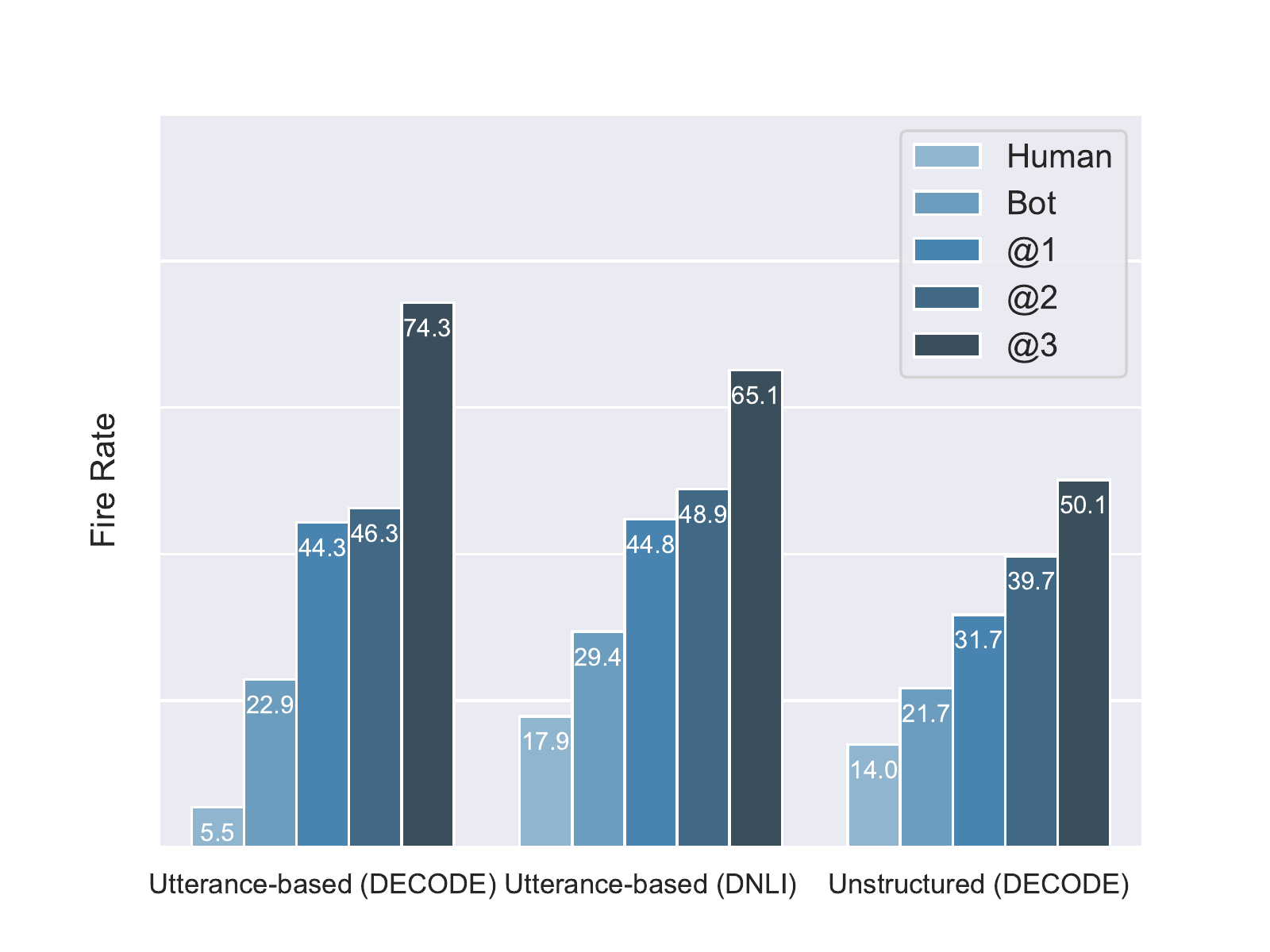}
    \caption{The fire rate of RoBERTa models with different setups on utterances belonging to different categories. ``Human" and ``Bot" stand for utterances by the human or the bot prospectively. ``@$N$" indicates the category where $N$ annotators agreed on the contradiction label. The x-axis indicates different approaches and the text in parentheses denotes the training data.}
    \label{fig:fire_rate_human_judgement}
\end{figure}

\paragraph{Model vs. Human Judgement}
To further understand the detector predictions and how well they might align with human judgements, we conduct the following experiment.
We first divide all the utterances into two categories based on whether they are generated by a human or a bot.
Then, the bot-generated utterances that have been marked by annotators as contradicting utterances are categorized into three sets based on the number of annotators that agree on the contradiction label. By design, the more annotators that agree on the contradiction label, the more plausible that it is a contradiction. 
We examine detector model fire rate on the utterances in the 5 different categories and results are shown in \autoref{fig:fire_rate_human_judgement}. 
The fire rate of utterance-based RoBERTa trained on DECODE on human utterances is 5.5\% contrasting to the 74.3\% on 3-agreed contradicting utterances, whereas the fire rates of unstructured RoBERTa on different categories are more clustered together. 
This finding demonstrates that all the models can discriminate between utterances with a distinct nature, and the model predictions are aligned with human judgments. Moreover, the fire rate of a strong discriminative detector could be a useful quantity to stratify utterances.

\paragraph{Using DECODE as an Automatic Metric}
\begin{figure}[ht]
    \centering
    \includegraphics[width=\columnwidth, trim=0 0 0 0, clip]{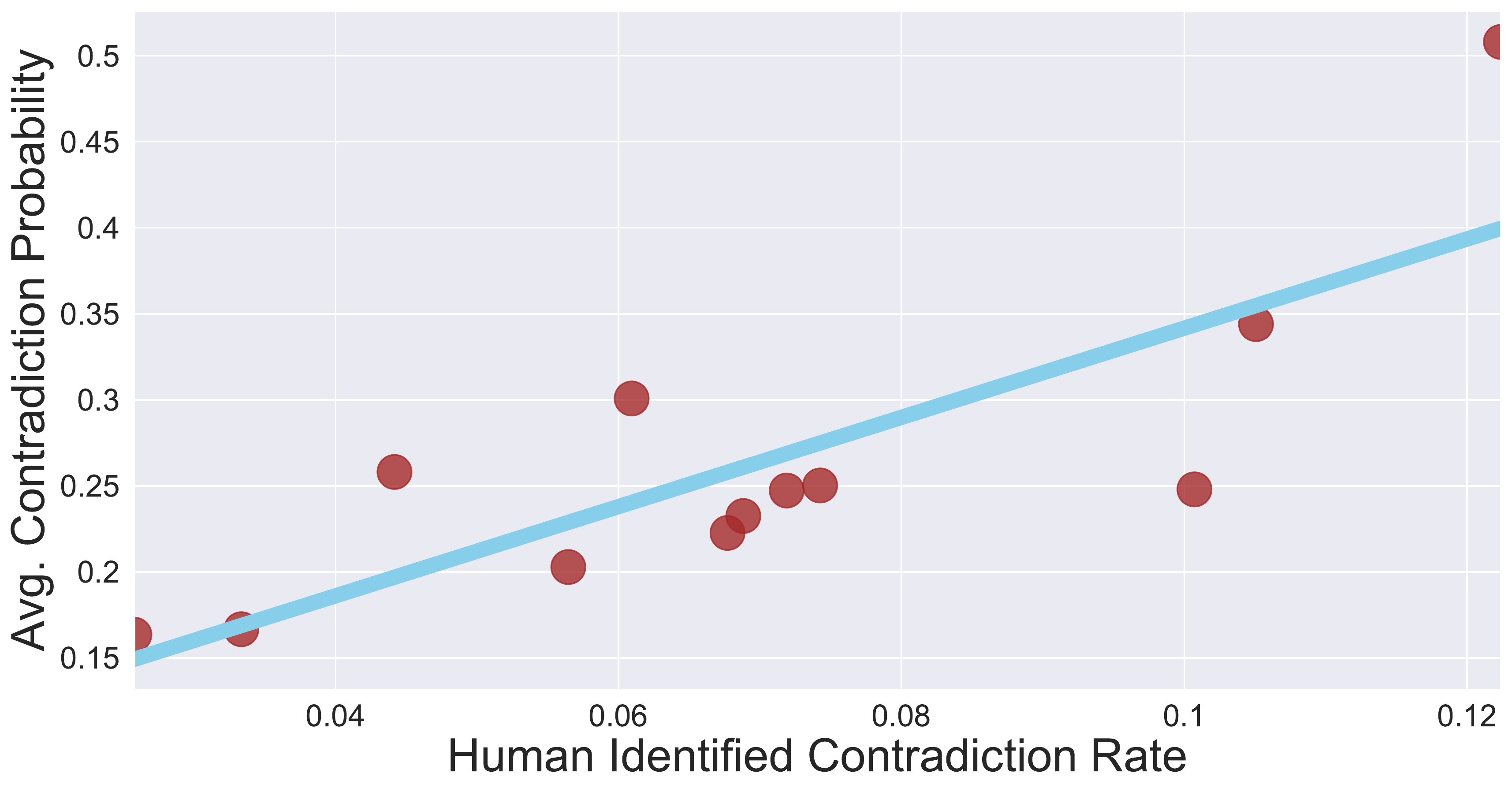}
    \caption{The comparison between the average contradiction score by the detector (y-axis) and the human identified contradiction rate (x-axis) on the utterances by different bots, averaged by type of bot. Each point in the plot is a bot which has conversed with humans and produced at least 180 utterances (with some identified as contradictions) in our interactive settings. The regression line shown yields a Pearson correlation coefficient of 0.81.}
    \label{fig:automatic metric}
\end{figure}
The results presented above indicate that the prediction of the detector can easily differentiate between the quality of utterances by humans and the utterances by bots. We further investigate whether it can differentiate the quality of the utterances by different bots and be used as an automatic metric checking generation consistency. We compare the average contradiction score of the detector with the contradiction rate by human judgements on the utterances generated by different classes of model (bots). The bots are the same set of models described in \autoref{sec:interactive} from which we collected our human-bot dialogue examples. The trend in \autoref{fig:automatic metric} reveals that the scores are positively correlated with human judgments, with a Pearson correlation coefficient of 0.81.
We would expect that improvement on the DECODE task will directly increase the correlation between the automatically produced detection score and human judgements, where use of such an automatic metric can ease the burden on laborious human evaluation of consistency.

\if 0
\begin{table}[t]
    \centering
    % \small
    \scalebox{0.88}{
    \begin{tabular}{lr}
    \toprule
\textbf{\small Model + Decoding Strategy} & \textbf{\small Contradict\%} \\

\midrule
%\multicolumn{2}{l}{\textit{Standard Generation Models}}\\
BST 2.7B Beam Search   &  38.1\% \\
BST 2.7B Top-$k$ ($k=40$) &  23.3\%\\
BST 2.7B Sample-and-Rank  &  29.6\%\\
\midrule
%\multicolumn{2}{l}{\textit{DECODE Re-ranked Models}}\\
% \hline
BST 2.7B Beam Search  + DECODE  &  22.7\% \\
BST 2.7B Top-$k$ ($k=40$) + DECODE &  1.1\% \\
\bottomrule
\end{tabular}
    }
    \caption{Generation Re-ranking using DECODE vs. standard methods, reporting the contradiction \% as flagged by our contradiction detection classifier (i.e., an automatic metric).}
    \label{tab:rerank_results}
\end{table}
\fi 

\begin{table}[t]
    \centering
    \small
    %\scalebox{0.88}{
    \begin{tabular}{lrr}
    \toprule
  \textbf{\small Model + }    & \multicolumn{1}{l}{\bf DECODE}  & \multicolumn{1}{l}{\bf Human}   \\
\textbf{\small Decoding Strategy} & \textbf{\small Contradict\%} & \textbf{\small Contradict\%} \\

\hline 
\midrule
{\em Standard generation} \\
%\hline 
%\multicolumn{2}{l}{\textit{Standard Generation Models}}\\
 Beam Search   &  38.1\%  & 38.3\%  \\
Top-$k$ ($k=40$) &  29.0\% & 31.8\%  \\
 Sample-and-Rank  &  29.6\% & 29.0\%   \\
\hline 
\midrule
{\em DECODE Re-ranking}\\
%\hline 
%\multicolumn{2}{l}{\textit{DECODE Re-ranked Models}}\\
% \hline
Beam Search   &  22.7\%  &  32.0\% \\
Top-$k$ ($k=40$)  &  1.1\% & 25.6\% \\
\bottomrule
\end{tabular}
    %}
    \caption{Generation Re-ranking using DECODE vs. standard methods, reporting the contradiction \% as flagged by our contradiction detection classifier (i.e., an automatic metric, ``DECODE Contradict\%'') in addition to  human judgments (``Human Contradict\%'').}
    \label{tab:rerank_results}
\end{table}

\subsection{Generation Re-ranking}

Given a contradiction detector, an obvious question other than using it as an automatic metric, is: can it be used to improve the consistency of dialogue generation models? We consider a very simple way to do that in the state-of-the-art generative model, BlenderBot (BST 2.7B)~\cite{roller2020recipes}.
During the decoding phase, for decoding methods that can output  multiple hypotheses, we simply rerank the top scoring hypotheses using the contradiction detection classifier.
We use our best performing classifier, our utterance-based RoBERTa model with DECODE fine-tuning, and consider three methods of decoding: beam search,  top-$k$ sampling~\cite{fan-etal-2018-hierarchical} and sample-and-rank~\cite{adiwardana2020towards},
and compare the standard and DECODE-reranked decoding methods to each other. For beam search we use the best found parameters from \cite{roller2020recipes} which are beam size 10, minimum beam length 20 and beam blocking of 3-grams. For top-$k$ we use $k=40$. For Sample-and-Rank we use 
$k$=40 and 20 samples. We consider the same human-bot dialogue logs as before, but only between Blenderbot BST 2.7B and humans, equally sampled between contradicting and non-contradicting utterances.
\autoref{tab:rerank_results} presents the results.

\paragraph{Automatic metric using DECODE}
Using our same DECODE contradiction classifier as the automatic metric, as in Sec. \ref{sec:interactive}. We observe that by re-ranking the beam of beam search (size 10) we can modestly improve the metric, but still 22.7\% of the time the detector flags generations as contradictions. Upon observation of the outputs, this appears to be because the beam of beam decoding tends to be not diverse enough \cite{vijayakumar2016diverse}, and when the top scoring utterance is flagged as contradicting, many of the other utterances in the beam are similar responses with slight rephrases, and are flagged contradicting as well.
Top-$k$ sampling fares much better, where reranking in our test can very often find at least one from the $k=40$ samples that does not flag the classifier, leaving only a 1.1\% contradiction firing rate. We note we expect these numbers are over-optimisticly low because the metric itself is being used to search (re-rank) and evaluate in this case.

\paragraph{Human Judgments}
The last column of \autoref{tab:rerank_results} presents human judgments of the various model generations,
judged using the same approach as before with three human verifiers, and reporting the percentage of contradictions.
%The results of this section are coming soon, but are currently still running. Hold tight, friends!
We observe similar results to the automatic metric findings: that DECODE re-ranking reduces the number of contradictions for both types of generation methods that we attempted to re-rank.

%While these are promising results, we note that while this paper deeply studies the {\em contradiction  detection} problem, here we have only scratched the surface of the  {\em non-contradiction generation} problem. Future work should address this further by studying and analysing the results of these techinques more deeply, as well as considering other methods than simply rescoring during decoding. 

\section{Conclusion}

We introduce the DialoguE COntradiction DEtection task (DECODE) and a new conversational dataset containing both human-human and human-bot contradictory dialogues. Training models on DECODE achieves better performance than other existing NLI data by a large margin. We further propose a structured utterance-based approach where each utterances are paired with other utterance before being fed into Transformer NLI models to tackle the dialogue contradiction detection task. We show the superiority of such an approach when transferring to out-of-distribution dialogues compared to a standard unstructured approach representative of mainstream NLU modeling. This is a valuable property since intermediate in-domain data are often scarce when integrating NLU module into NLG systems. We further show that our best contradiction detector correlates with human judgements, and provide evidence for its usage in both automatic checking and improving the consistency of state-of-the-art generative chatbots.

While this paper deeply studies the {\em contradiction  detection} problem, we believe here we have only scratched the surface of the  {\em non-contradiction generation} problem, while obtaining promising first results in that setting. Future work should address this further by studying and analysing the results of these techniques more deeply, as well as considering other methods than simply rescoring during decoding. 
Going forward, we envision complementary progress on both the modeling of NLU and NLG and the integration of the two. We hope our work could facilitate and provide guidelines for future work on incorporating NLU modeling into dialogue systems.

% NLG is in a new era where the thing we care about can not be esily evaluated through automatic lexical related metrics like ppl.
% 
% intermediate , progress on NLU is on track.
% We envision complementary progress on the modeling of NLU and NLG where NLU module.  

% main result in the table
% auxiliary results shown in a bar.
% organize better here for other stuff.

\bibliography{anthology,ref}
\bibliographystyle{acl_natbib}

\clearpage

\appendix

\section{Verification Statistics}
\label{sec:verification_statistics}
For a subset of the contradicting dialogues in DECODE  we asked three verifiers to determine whether the original writer indeed created a contradiction example. \autoref{tab:verification_statistics} shows the verification statistics. Note that we only use examples on which all three verifiers agreed for DECODE (dev) and DECODE (Test).

\begin{table}[t]
    \centering
    % \small
    \scalebox{1}{
    \begin{tabular}{crr}
    \toprule
    \textbf{\# of Verifiers Agreed} & \textbf{Count} & \textbf{Ratio (\%)} \\
    \midrule
    0 & 484 & 7.67\% \\
    1 & 497 & 7.87\% \\
    2 & 1,211 & 19.18\% \\
    3 & 6,214 & 65.28\%\\
    \bottomrule
    \end{tabular}}
    \caption{Verification Statistics. The first column indicates the number of verifiers that agreed upon the  given contradictions.}
    \label{tab:verification_statistics}
\end{table}

\end{document}